\documentclass{article} 
\usepackage{iclr2015}
\usepackage{hyperref}
\usepackage{url}
\usepackage{graphicx}
\usepackage{amsmath,amsfonts,amssymb}
\usepackage{soul}
\usepackage{natbib}
\usepackage{hyperref}

\usepackage{caption}
\definecolor{lightgreen}{rgb}{.5,1,0}
\sethlcolor{lightgreen}
\newcommand{\highlight}[1]{\hl{\textbf{#1}}}

\title{Extraction of Salient Sentences from\\Labelled Documents}

\author{
Misha Denil$^1$ \quad Alban Demiraj$^1$ \quad
\textbf{Nando de Freitas}$^{1,2,3}$ \\
$^1$University of Oxford, United Kingdom\\
$^2$Canadian Institute for Advanced Research\\
$^3$Google DeepMind\\
\texttt{misha.denil@cs.ox.ac.uk}\\
\texttt{a.demiraj@oxfordalumni.org}\\
\texttt{nandodefreitas@google.com}}

\iclrconference

\newcommand{\vS}{\mathbf{S}}
\newcommand{\vw}{\mathbf{w}}
\newcommand{\vW}{\mathbf{W}}
\newcommand{\vf}{\mathbf{f}}
\newcommand{\vF}{\mathbf{F}}

\newcommand{\R}{\mathbb{R}}

\begin{document}

\maketitle

\begin{abstract}
We present a hierarchical convolutional document model with an architecture designed to support introspection of the document structure.  Using this model, we show how to use visualisation techniques from the computer vision literature to identify and extract topic-relevant sentences.

We also introduce a new scalable evaluation technique for automatic sentence extraction systems that avoids the need for time consuming human annotation of validation data.
\end{abstract}


\section{Introduction}

The idea of encoding symbolic concepts with distributed representations has excited researchers for decades \citep{Hinton1986,Bengio2003}.  In recent years this idea has re-emerged following the success of neural networks in many natural language processing domains such as language modelling \citep{Mikolov2013a}, machine translation \citep{Schwenk2012,Kalchbrenner2013a,Devlin2014,cho2014,Bahdanau2014,Sutskever2014,Hermann:2014:ACLphil}, question answering \citep{BordesCW14,WestonCB14}, dialogue systems \citep{kalchbrenner2013recurrent}, sentiment analysis \citep{Socher2011,Socher2012,Hermann2013a}, and other natural language processing tasks such as chunking and named entity recognition \citep{Collobert2011}.  

Distributed representations of language enable us to encode the semantics of language into the geometry of a continuous vector space.  This idea underlies nearly all modern applications of neural networks to NLP.  One intuition behind the success of continuous representations is that they are more amenable to transformations and optimizations than their discrete counterparts \citep{bottou-mlj-2013}.

Following their tremendous success in computer vision, there have recently been many successful applications of Convolutional neural networks (ConvNets for short) to NLP \citep{Collobert2011,NalKalchbrennerGrefenstette2014,Hermann:2014:ACLphil,DosSantos2014,Hu2014,Yu2014}.

In this paper we make two main contributions, continuing the line of research on ConvNets for language:
\begin{enumerate}
\item We introduce a hierarchical ConvNet architecture designed to support introspection of the document structure.  We show how our novel model structure allows us to adapt visualisation techniques from computer vision to design a system that can automatically extract relevant sentences from labelled documents.
\item We describe a new evaluation technique for automatic sentences extraction systems that can be applied at scale.  We can take advantage of the fact that the documents are labelled to build a simple document classifier which we then use to classify the documents extracted by our system.  Comparing the accuracy of the classifier on full documents to its accuracy on the extracted portions alone gives a measure of how much task-relevant information is preserved by the extraction process.
\end{enumerate}

Extracting relevant sentences from documents is an important problem.  Summaries created in this way pervade internet search results, restaurant and consumer product review sites and the front pages of news outlets.  Our work shows a novel application of neural networks to this problem, and provides a scalable method for comparing the quality of different solutions.

\section{Model description}
\label{sec:model-description}

Our ConvNet model is divided into a sentence level and a document level, both of which are implemented using convolutions.  At the sentence level we use a ConvNet to transform embeddings for the words in each sentence into an embedding for the entire sentence.  At the document level we use another ConvNet to transform sentence embeddings from the first level into a single embedding vector that represents the entire document.

The model is trained by feeding the document embeddings into a softmax classifier, and the model is trained by backpropogation through both the sentence and document levels.  At the sentence level the filter banks that process different sentences are tied, so each sentence has an embedding that is produced by the same ConvNet.

The architecture of our model forces information to pass through an intermediate sentence based representation.  This architecture is inspired by \cite{Gulcehre2013} who show that learning appropriate intermediate representations helps generalisation, and also by \cite{Hinton2011} who show that by forcing information to pass through carefully chosen bottlenecks it is possible to control the types of intermediate representations that are learned.

Forcing the model to represent documents in terms of semantically relevant units is essential to the extraction application we introduce later in this paper.  The fact that specific hidden units in our network are used to represent each sentence is what allows us to measure the relevance contribution from each sentence separately.

Each level of our model in isolation is very similar to the DCNN of \citet{NalKalchbrennerGrefenstette2014}; the main difference is in how interactions between embedding dimensions are handled.  Our approach (explained below) follows previous work on ConvNets for NLP (e.g. \citet{Collobert2011}) and is also consistent with ConvNets for vision.  Using a more standard form of convolution makes the model more compact (which is nice when adding more layers) but is not the focus of this paper.  The main difference between our model and the DCNN is that we extend the model with a multi-level structure for modelling documents, where \citet{NalKalchbrennerGrefenstette2014} considered only sentences.

\begin{figure}[t]
\includegraphics[width=1.0\linewidth]{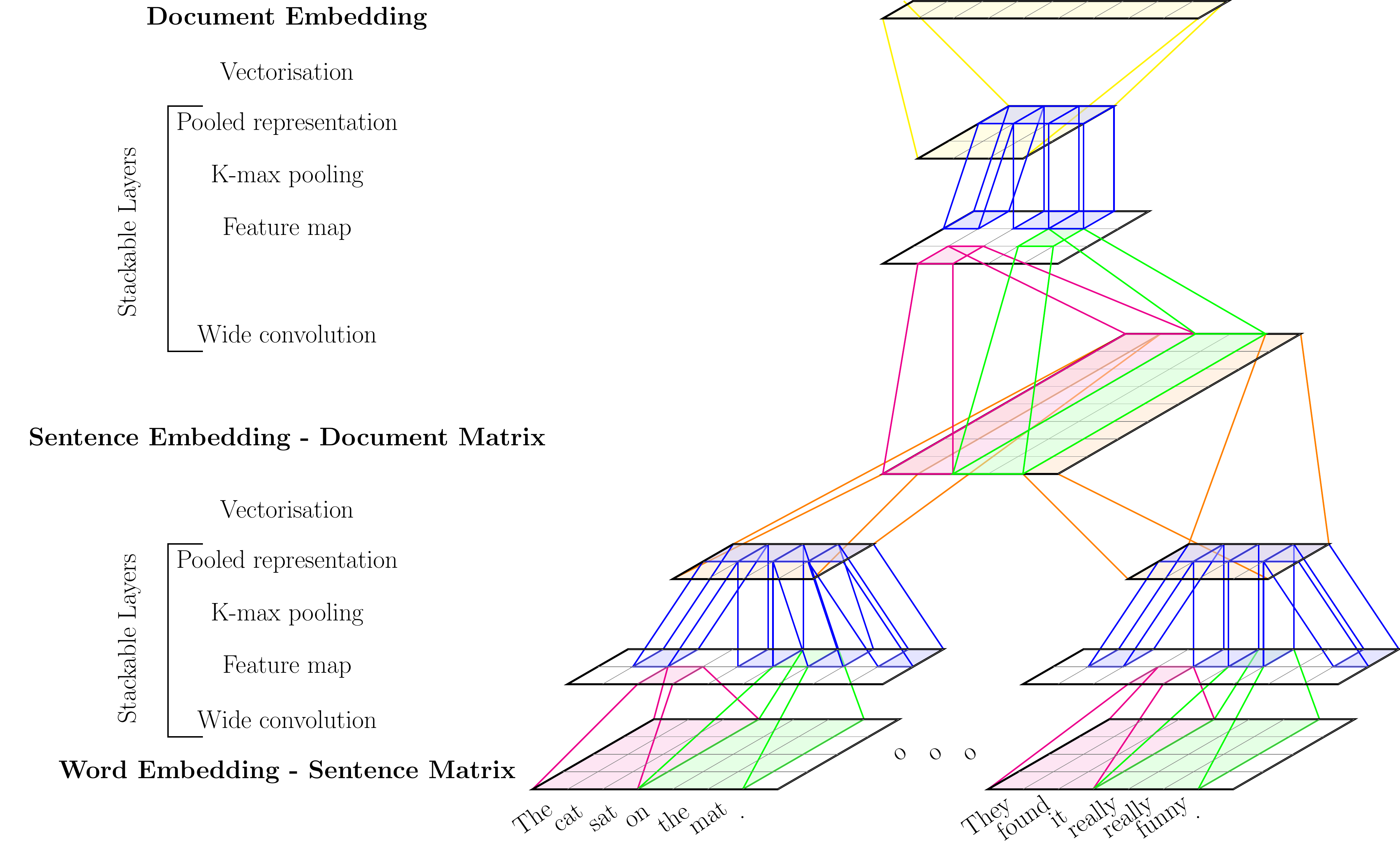}
\vspace{10pt}
\caption{Word embeddings are concatenated into columns to form a sentence matrix.  The sentence level ConvNet applies a cascade of convolution, pooling and nonlinearity operations to transform the projected sentence matrix into an embedding for the sentence.  The sentence embeddings are then concatenated into columns to form a document matrix.  The document model then applies its own cascade of convolution, pooling and nonlinearity operations to form an embedding for the whole document which is fed into a softmax layer for classification.}
\label{fig:cdm}
\end{figure}

A detailed schematic of our model is shown in Figure~\ref{fig:cdm}.  In the following sections we describe the inner workings of each part in more detail.\footnote{Code implementing this model will be made available following the review process.}

\subsection{Embedding matrix}

The input to each level of our model is an embedding matrix.  At the sentence level the columns of this matrix correspond to embeddings of the words in the sentence being processed, while at the document level the columns correspond to sentence embeddings produced by the sentence level of the model.

At the sentence level, an embedding matrix for each sentence is built by concatenating embeddings for each word it contains into the columns of a matrix.  The words are drawn from a fixed vocabulary $V$, which we represent using a matrix of word embeddings $\vW \in \R^{d\times |V|}$. Each column of this matrix is a $d$ dimensional vector that gives an embedding for a single word in the vocabulary.  The word embedding vectors are parameters of the model, and are optimised using backpropogation.

\begin{align*}
\vW = \begin{bmatrix}
| & | & | \\
\vw_{1} &  \cdots & \vw_{|V|} \\
| & | & |
\end{bmatrix}
\end{align*}

For each sentence $s = \begin{bmatrix} w^s_1 & \cdots & w^{s}_{|s|} \end{bmatrix}$ we generate a sentence matrix $\vS_s \in \R^{d\times|s|}$ by concatenating together the word embedding vector corresponding to each word in the sentence.  

\begin{align*}
\vS_s = \begin{bmatrix}
| & | & | \\
\vw_{w^s_1} &  \cdots & \vw_{w^s_{|s|}} \\
| & | & |
\end{bmatrix}
\end{align*}

The sentence level of the model produces an embedding vector for each sentence in the document.  The input to the document level is obtained by assembling these sentence embeddings into a document matrix, in the same way word embeddings are assembled into a sentence matrix in the level below.

\subsection{Convolution}

Each convolutional layer contains a filter bank $\vF \in \R^{d\times w_f \times n_f}$ where $w_f$ and $n_f$ refer to the width and number of feature maps respectively.  The first dimension of each feature map $\vf \in \R^{d\times w_f}$ is equal to the number of dimensions in the embeddings generated by the layer below.

The convolution operation in our model is one dimensional.  We align the first axis of each feature map with the embedding axis and convolve along rows of the embedding matrix.  At the sentence level this corresponds to convolving across words and at the document level it corresponds to convolving across sentences.

Each feature map generates a 1d row of numbers, where each value is obtained by applying the feature map at a different location along the sentence matrix.  The outputs of different feature maps are then stacked to form a new matrix of latent representations which is fed as input to the next layer.  In all cases we use wide (``\texttt{full}'') convolutions in order for all weights in the feature maps to reach every word/sentence, including ones on the edges.



\subsection{k-max pooling}

Since different sentences and documents have different lengths, not all embedding matrices will be of the same width.  This is not an issue for the convolutional layers, since convolutions can handle inputs of arbitrary width, but is problematic to use as input to a fully connected or softmax layer, and at the interface between the sentence and document levels.

The solution we use is $k$-max pooling, which is applied to each row of the embedding matrix separately.  To apply $k$-max pooling to a single row, we keep the $k$ largest values along that row and discard the rest.  Since $k$ is a fixed parameter this always generates a fixed size output (if the input has length less than $k$ we pad it with zeros).  For example, applying 2-max pooling to $[3,1,5,2]$ yields $[3,5]$.  This procedure is also illustrated graphically in Figure~\ref{fig:cdm}.

\subsection{Full model}

At the sentence level we use a single convolutional pipeline for processing each sentence (ie.\ the weights of the sentence models are tied across sentences).  The document level of the model is composed of the same type of convolution and pooling primitives, but the weights are not shared with the sentence level models.

An important feature of this archetecture is that it forces the model to create an intermediate representation for each sentence in the document.  It is this representation that allows the extraction technique in Section~\ref{sec:visualization} to identify salient sentences.  If we had instead simply applied convolutions to the whole document we would not be able to disentangle the contributions from individual sentences.


\section{Sentence extraction through visualisation}
\label{sec:visualization}

Deconvolutional networks~\citep{Zeiler2010,Zeiler2011} have been used to great effect to generate interpretable visualisations of the activations in deep layers of convolutional neural networks in computer vision~\citep{Zeiler2012}.
More recent work has shown that good visualisations can be obtained by using a single backpropogation pass through the network.  In fact this procedure is formally quite similar to the operations carried out in a deconvolutional net~\citep{Simonyan2013}.  Visualisation through backpropogation is a generalisation of the deconvolutional approach, since one can backpropogate through non-convolutional layers.

The first step in our extraction procedure is to create a saliency map for the document by assigning an importance score to each sentence.  To generate the saliency map for a given document, we adopt the technique of \cite{Simonyan2013} with a modified objective function.

We first perform a forward pass through the network to generate a class prediction for the document.  We then construct a \emph{pseudo-label} by inverting the network predictions, and feed this to the loss function as the true label.  This choice of pseudo-label allows us to induce the greatest loss.

To infer saliency for words we take a first order Taylor expansion of the loss function using the pseudo-label.  Formally, if $x$ is a document and $f(x)$ is the network evaluated on $x$, we approximate the loss as a linear function
\begin{align*}
L(\tilde{y}, f(x)) \approx w^T x + b
\end{align*}
where $\tilde{y}$ is the inverted label and
\begin{align*}
w = \frac{\partial L}{\partial x}\bigg|_{(\tilde{y}, f(x))} \enspace.
\end{align*}
The vector $w$ has one entry for each word in the document and we can use $|w_i|$ as a measure of the saliency of word $i$.  These saliency scores can be easily computed by performing a single pass of backpropogation through the network.  The intuition behind using gradient magnitudes as a saliency measure is that the magnitude of the derivative indicates which words need to be changed the least to affect the score the most.

When training the model we need only backpropogate to the word embedding layer to get a gradient with respect to each word embedding vector; however, to evaluate saliency we need a derivative with respect to the words themselves.  To bacpropogate through the embedding layer, notice that (using the notation of Section~\ref{sec:model-description}) we can write the sentence matrix as
\begin{align*}
\mathbf{S}_s = \mathbf{W}\mathbf{I}_s
\end{align*}
where $\mathbf{I}_s \in \mathbb{R}^{|V|\times|s|}$ is a matrix whose columns are 1-hot vectors identifying the dictionary index of each word in the sentence $s$.  If the $i$th word in $s$ appears at index $j$ in the dictionary then the derivative with respect to $I_{ji}$ is given by
\begin{align*}
w_i = \frac{\partial L}{\partial I_{ji}}\bigg|_{(\tilde{y}, f(x))} = \langle \boldsymbol{\delta}_i, \mathbf{w}_j \rangle
\end{align*}
which corresponds to taking the dot product between $\boldsymbol{\delta}_i$, the backwards message for the $i$th embedding vector; and $\mathbf{w}_i$, the embedding vector for the $i$th word.

At the document level the situation is similar, except that the dictionary of sentence embeddings is defined implicitly by the sentence level of our model.  Using $\mathbf{S}$ to denote the (huge) implicitly defined dictionary of embeddings that contains a column for every possible sentence we can write down the first layer of the of the document level of our model as
\begin{align*}
\mathbf{D}_d = \mathbf{S}\mathbf{I}_d
\end{align*}
where $\mathbf{D}_d$ is the document matrix and $\mathbf{I}_d$ is a matrix whose columns are (huge) 1-hot vectors indicating the dictionary index of each sentence in the document $d$.  If the $i$th sentence in $d$ appears at index $j$ in the implicit dictionary then the derivative with respect to $I_{ji}$ is given by 
\begin{align*}
w_i = \frac{\partial L}{\partial I_{ji}}\bigg|_{(\tilde{y}, f(x))} = \langle \boldsymbol{\delta}_i, \mathbf{s}_j \rangle = \langle \boldsymbol{\delta}_i, f_s(x_j) \rangle
\end{align*}
where $\boldsymbol{\delta}_i$ is again the backward message from backpropogation and $\mathbf{s}_j = f_s(x_j)$ is the result of evaluating the sentence level of the model on sentence $j$.


We use the saliency scores for each sentence to rank the sentences within a document.  To extract a fixed number of sentences we simply choose the $k$ most highly ranked sentences using this measure.

\section{Scalable evaluation}
\label{sec:evaluation}

The sentence extraction scheme described in Section~\ref{sec:visualization} qualitatively produces very good results (see Figure~\ref{fig:example-summaries} for several examples); however, we would also like a quantitative measure of performance that allows us to compare different extraction methods.  An obvious choice for evaluation would be metrics used in summarisation; however, since we lack labels for ``correct'' extractions we do not have a gold standard to compare against.

Instead, we propose a way to measure extraction quality that can be easily applied to labelled documents at scale.  We propose to train a simple ``reference'' model on full documents, and then compare the performance of the reference model on full documents to its performance on documents created by extracting a small number of sentences. 

This evaluation scheme is somewhat unorthodox, but comes with a very intuitive interpretation.  If the extraction process chooses task relevant sentences then it should be easy for the reference model to identify the label of the original document based on an extracted subset of sentences.  On the other hand, choosing irrelevant sentences should confuse the reference model, reducing its accuracy.

\section{Experiments}

\subsection{Data preparation}

We use the IMDB movie review sentiment data set, which was originally introduced by \cite{Maas2011}, to demonstrate our extraction technique.  This dataset contains a total of 100000 movie reviews posted on IMDB.  There are 50000 unlabelled reviews and the remaining 50000 labelled reviews are divided into a 25000 review training set and a 25000 review test set.  Each of the labelled reviews has a binary label, either positive or negative.  In our experiments, we use only the labelled portion of this data set.

We pre-process each review by first stripping HTML markup and breaking the review into sentences and then breaking each sentence into words.  We use NLTK\footnote{\url{http://www.nltk.org/}} to perform these tasks.  We also map numbers to a generic \texttt{NUMBER} token, any symbol that is not in \texttt{.?!} to \texttt{SYMBOL} and any word that appears fewer than 5 times in the training set to \texttt{UNKNOWN}.  This leaves us with a 29493 word vocabulary.

\subsection{Review Summarisation via Sentence extraction}

\begin{table}
\begin{center}
\begin{tabular}{lrrr|lrrr}
Proportion & ConvNet & Word2Vec & Rand. & Fixed & ConvNet & Word2Vec & Rand. \\ \hline\hline
 50\% & \textbf{82.74} & 81.98 & 79.79 & Pick 5 & \textbf{83.12} & 82.26 & 80.02 \\
 33\% & \textbf{82.72} & 80.39 & 76.72 & Pick 4 & \textbf{82.91} & 81.92 & 79.05 \\
 25\% & \textbf{82.94} & 80.18 & 74.87 & Pick 3 & \textbf{82.59} & 81.48 & 77.15 \\
 20\% & \textbf{82.84} & 79.70 & 73.20 & Pick 2 & \textbf{81.71} & 80.39 & 74.48 \\ \hline
Full & 83.04 \\
First+Last & 68.62
\end{tabular}
\end{center}
\caption{Results of classifying extracted documents with the reference model.  Results labelled \emph{Proportion} indicate selecting up to the indicated percentage of sentences in the review, and results labelled \emph{Fixed} show the result of selecting a fixed number of sentences from each.  \emph{ConvNet} shows reference model performance on extractions produced by our ConvNet.  \emph{Word2Vec} shows reference performance on extractions produced by the shallow baseline. \emph{Rand} shows reference performance on randomly created extractions.  The final two rows show reference accuracy on the full reviews (\emph{Full}) and when using the common heuristic of extracting the first and last sentence (\emph{First+Last}).}
\label{tab:naive-bayes}
\end{table}


\begin{figure}[p]
\small{I caught this movie on the Sci-Fi channel recently. It actually turned out to be pretty decent as far as B-list horror/suspense films go. \highlight{Two guys (one naive and one loud mouthed a **) take a road trip to stop a wedding but have the worst possible luck when a maniac in a freaky, make-shift tank/truck hybrid decides to play cat-and-mouse with them.} Things are further complicated when they pick up a ridiculously whorish hitchhiker. What makes this film unique is that the combination of comedy and terror actually work in this movie, unlike so many others. The two guys are likable enough and there are some good chase/suspense scenes. Nice pacing and comic timing make this movie more than passable for the horror/slasher buff. \highlight{Definitely worth checking out.}}
\\[0.2cm]
\small{I just saw this on a local independent station in the New York City area. \highlight{The cast showed promise but when I saw the director, George Cosmotos, I became suspicious. And sure enough, it was every bit as bad, every bit as pointless and stupid as every George Cosmotos movie I ever saw.} He's like a stupid man's Michael Bey -- with all the awfulness that accolade promises. There's no point to the conspiracy, no burning issues that urge the conspirators on. We are left to ourselves to connect the dots from one bit of graffiti on various walls in the film to the next. Thus, the current budget crisis, the war in Iraq, Islamic extremism, the fate of social security, 47 million Americans without health care, stagnating wages, and the death of the middle class are all subsumed by the sheer terror of graffiti. A truly, stunningly idiotic film.}
\\[0.2cm]
\small{Graphics is far from the best part of the game. \highlight{This is the number one best TH game in the series.} Next to Underground. \highlight{It deserves strong love. It is an insane game.} There are massive levels, massive unlockable characters... it's just a massive game. \highlight{Waste your money on this game. This is the kind of money that is wasted properly.} And even though graphics suck, thats doesn't make a game good. Actually, the graphics were good at the time. Today the graphics are crap. WHO CARES? As they say in Canada, This is the fun game, aye. (You get to go to Canada in THPS3) Well, I don't know if they say that, but they might. who knows. Well, Canadian people do. Wait a minute, I'm getting off topic. This game rocks. Buy it, play it, enjoy it, love it. It's PURE BRILLIANCE.}
\\[0.2cm]
\small{The first was good and original. I was a not bad horror/comedy movie. So I heard a second one was made and I had to watch it . What really makes this movie work is Judd Nelson's character and the sometimes clever script. \highlight{A pretty good script for a person who wrote the Final Destination films and the direction was okay.} Sometimes there's scenes where it looks like it was filmed using a home video camera with a grainy - look. Great made - for - TV movie. \highlight{It was worth the rental and probably worth buying just to get that nice eerie feeling and watch Judd Nelson's Stanley doing what he does best.} I suggest newcomers to watch the first one before watching the sequel, just so you'll have an idea what Stanley is like and get a little history background.}
\\[0.2cm]
\small{Why do all movies on Lifetime have such anemic titles? \highlight{``An Unexpected Love'' - ooh, how provocative!!} ``This Much I know'' would have been better. \highlight{The film is nothing special.} Real people don't really talk like these characters do and the situations are really hackneyed. The straight woman who ``turns'' lesbian seemed more butch than the lesbian character. If you wanna watch two hot women kiss in a very discreet fashion, you might enjoy this. Although it seems like it was written by someone who doesn't really get out in the world to observe people. Why am I wasting my time writing about it?}
\\[0.2cm]
\small{A friend and I went through a phase some (alot of) years ago of selecting the crappest horror films in the video shop for an evening's entertainment. For some reason, I ended up buying this one (probably v. v. cheap). \highlight{The cheap synth soundtrack is a classic of its time and genre.} There's also a few very amusing scenes. Among them is a scene where a man's being attacked and defends himself with a number of unlikely objects, it made me laugh at the time (doesn't seem quite so funny in retrospect but there you go). \highlight{Apart from that it's total crap, mind you.} But probably worth a watch if you like films like ``Chopping Mall''. Yes, I've seen that too.}
\\[0.2cm]
\small{Vertigo co - stars Stewart (in his last turn as a romantic lead) and Novak elevate this, Stewart's other ``Christmas movie,'' movie to above mid - level entertainment. \highlight{The chemistry between the two stars makes for a fairly moving experience and further revelation can be gleaned from the movie if witchcraft is seen as a metaphor for the private pain that hampers many people's relationships.} All in all, a nice diversion with legendary stars, 7/10}
\\[0.2cm]
\caption{Several example extractions chosen by our ConvNet.  The full text of the review is shown in black and the sentences selected by the ConvNet appear in colour. While choosing the first and last sentence is a popular pragmatic approach, it is clear from these examples that this heuristic is not as effective as our ConvNet based scheme.  In each example we select up to 20\% of the sentences in the review for extraction.}
\label{fig:example-summaries}
\end{figure}

We train our ConvNet model on the IMDB movie review data set.  The sentence level model uses 10-dimensional word embeddings which are convolved with 6 feature maps of width 5, followed by a 4-max pooling layer and a tanh nonlinearity.  The weights of this model are tied across sentences in a document.  The document level model convolves its input with a bank of 15 feature maps of width 5, followed by 2-max pooling and a tanh nonlinearity.  Finally the result is fed into a softmax classifier which predicts if the sentiment of the review is positive or negative.

We extract salient sentences from each of the reviews in the IMDB movie review test set using the method of Section~\ref{sec:visualization}.  For evaluation, we also train a reference model using Na\"ive Bayes on the full training set.

We compare the performance of our model against several baselines.  
\begin{enumerate}
\item A shallow neural network model with a single hidden layer.  The model uses word2vec to obtain word embeddings, which it then combines into document embeddings by summing the words in the document.  It then applies a logistic regression to the document embedding to predict the sentiment of the document.  We apply the visualisation technique of Section~\ref{sec:visualization} to choose sentences to extract in the same way as for the ConvNet model.
\item A random model assigns relevance scores to sentences uniformly at random.
\item A ``First+last'' model which follows the common heuristic that the first and last sentence of a document are often very informative about its content.
\end{enumerate}

The results of this experiment are shown in Table~\ref{tab:naive-bayes}.   Even keeping only two sentences from each review, the accuracy of the reference model drops by less than 1.5\% on the test set.  In all cases our convolutional model outperforms all of the baselines we compare against.

We show several examples of extractions created by our model in Figure~\ref{fig:example-summaries}.  As can be seen, many of the reviews begin with short descriptions of where the reviewer saw the film, or with a brief summary of the plot.  These sentences are not useful since they do not express an opinion on the movie being reviewed.  Our model learns to ignore these background sentences very consistently.


\section{Conclusion}

In this paper we introduced a ConvNet model for documents with an architecture designed to support introspection of the document structure.  We have also shown that we can apply the visualisation technique of \cite{Simonyan2013} to identify and extract task-specific salient sentences from documents.  We demonstrated this technique by extracting sentiment-relevant sentences from movie reviews.

We also introduced a scalable evaluation method for automatic sentence extraction systems.  By comparing the performance of a reference classifier on full documents to its performance on documents created from extracted sentences we can gauge how much task-relevant information is preserved by the extraction process.

We compared our ConvNet extraction model to several baseline extraction methods.  Our evaluation shows that our model extracts more task-relevant information than the baseline methods, even when the total number of extracted sentences is very small.

\subsubsection*{Acknowledgments}
We would like to thank Phil Blunsom and Nal Kalchbrenner for many interesting discussions about ConvNets for NLP and for their contributions to an early version of this work.

\small{
\bibliography{txtnets}

\begin{thebibliography}{29}
\providecommand{\natexlab}[1]{#1}
\providecommand{\url}[1]{\texttt{#1}}
\expandafter\ifx\csname urlstyle\endcsname\relax
  \providecommand{\doi}[1]{doi: #1}\else
  \providecommand{\doi}{doi: \begingroup \urlstyle{rm}\Url}\fi

\bibitem[Bahdanau et~al.(2014)Bahdanau, Cho, and Bengio]{Bahdanau2014}
Bahdanau, Dzmitry, Cho, Kyunghyun, and Bengio, Yoshua.
\newblock Neural machine translation by jointly learning to align and
  translate.
\newblock Technical report, University of Montreal, 2014.

\bibitem[Bengio et~al.(2003)Bengio, Ducharme, Vincent, and Jauvin]{Bengio2003}
Bengio, Yoshua, Ducharme, Rejean, Vincent, Pascal, and Jauvin, Christian.
\newblock {A Neural Probabilistic Language Model}.
\newblock \emph{Journal of Machine Learning Research}, 3:\penalty0 1137--1155,
  2003.

\bibitem[Bordes et~al.(2014)Bordes, Chopra, and Weston]{BordesCW14}
Bordes, Antoine, Chopra, Sumit, and Weston, Jason.
\newblock Question answering with subgraph embeddings.
\newblock \emph{CoRR}, abs/1406.3676, 2014.

\bibitem[Bottou(2014)]{bottou-mlj-2013}
Bottou, L\'{eon}.
\newblock From machine learning to machine reasoning: an essay.
\newblock \emph{Machine Learning}, 94:\penalty0 133--149, January 2014.

\bibitem[Cho et~al.(2014)Cho, van Merrienboer, Gulcehre, Bougares, Schwenk, and
  Bengio]{cho2014}
Cho, Kyunghyun, van Merrienboer, Bart, Gulcehre, Caglar, Bougares, Fethi,
  Schwenk, Holger, and Bengio, Yoshua.
\newblock Learning phrase representations using rnn encoder-decoder for
  statistical machine translation.
\newblock 2014.

\bibitem[Collobert et~al.(2011)Collobert, Weston, Bottou, Karlen, Kavukcuoglu,
  and Kuksa]{Collobert2011}
Collobert, R, Weston, Jason, Bottou, Leon, Karlen, Michael, Kavukcuoglu, Koray,
  and Kuksa, Pabvel.
\newblock {Natural language processing (almost) from scratch}.
\newblock \emph{The Journal of Machine Learning Research}, 12:\penalty0
  2461--2505, 2011.

\bibitem[Devlin et~al.(2014)Devlin, Zbib, Huang, Lamar, Schwartz, and
  Makhoul]{Devlin2014}
Devlin, Jacob, Zbib, Rabih, Huang, Zhongqiang, Lamar, Thomas, Schwartz,
  Richard, and Makhoul, John.
\newblock {Fast and Robust Neural Network Joint Models for Statistical Machine
  Translation}.
\newblock In \emph{Association for Computational Linguistics}, 2014.

\bibitem[dos Santos \& Gatti(2014)dos Santos and Gatti]{DosSantos2014}
dos Santos, Ciccero~Noguelra and Gatti, Maira.
\newblock {Deep convolutional neural networks for sentiment analysis of short
  texts}.
\newblock In \emph{International Conference on Computational Linguistics},
  2014.

\bibitem[Gulcehre \& Bengio(2013)Gulcehre and Bengio]{Gulcehre2013}
Gulcehre, Caglar and Bengio, Yoshua.
\newblock {Knowledge Matters: Importance of Prior Information for
  Optimization}.
\newblock In \emph{International Conference on Learning Representations}, 2013.

\bibitem[Hermann \& Blunsom(2013)Hermann and Blunsom]{Hermann2013a}
Hermann, Karl~Moritz and Blunsom, Phil.
\newblock The role of syntax in vector space models of compositional semantics.
\newblock \emph{Proceedings of the ACL}, pp.\  894--904, 2013.

\bibitem[Hermann \& Blunsom(2014)Hermann and Blunsom]{Hermann:2014:ACLphil}
Hermann, Karl~Moritz and Blunsom, Phil.
\newblock {Multilingual Models for Compositional Distributional Semantics}.
\newblock In \emph{Proceedings of ACL}, 2014.

\bibitem[Hinton et~al.(2011)Hinton, Krizhevsky, and Wang]{Hinton2011}
Hinton, G~E, Krizhevsky, A, and Wang, S~D.
\newblock {Transforming Auto-encoders}.
\newblock In \emph{International Conference on Artificial Neural Networks},
  2011.

\bibitem[Hinton(1986)]{Hinton1986}
Hinton, Geoffrey~E.
\newblock {Learning Distributed Representations of Concepts}.
\newblock In \emph{Annual Conference of the Cognitive Science Society}, pp.\
  1--12, 1986.

\bibitem[Hu et~al.(2014)Hu, Lu, Li, and Chen]{Hu2014}
Hu, Baotian, Lu, Zhengdong, Li, Hang, and Chen, Qingcai.
\newblock Convolutional neural network architectures for matching natural
  language sentences.
\newblock In \emph{Advances in Neural Information Processing Systems 27}, pp.\
  2042--2050, 2014.

\bibitem[Kalchbrenner \& Blunsom(2013{\natexlab{a}})Kalchbrenner and
  Blunsom]{Kalchbrenner2013a}
Kalchbrenner, Nal and Blunsom, Phil.
\newblock {Recurrent Continuous Translation Models}.
\newblock In \emph{Empirical Methods in Natural Language Processing},
  2013{\natexlab{a}}.

\bibitem[Kalchbrenner \& Blunsom(2013{\natexlab{b}})Kalchbrenner and
  Blunsom]{kalchbrenner2013recurrent}
Kalchbrenner, Nal and Blunsom, Phil.
\newblock Recurrent convolutional neural networks for discourse
  compositionality.
\newblock \emph{Proceedings of the 2013 Workshop on Continuous Vector Space
  Models and their Compositionality}, 2013{\natexlab{b}}.

\bibitem[Kalchbrenner et~al.(2014)Kalchbrenner, Grefenstette, and
  Blunsom]{NalKalchbrennerGrefenstette2014}
Kalchbrenner, Nal, Grefenstette, Edward, and Blunsom, Phil.
\newblock {A Convolutional Neural Network for Modelling Sentences}.
\newblock In \emph{Association for Computational Linguistics}, 2014.

\bibitem[Maas et~al.(2011)Maas, Daly, Pham, Huang, Ng, and Potts]{Maas2011}
Maas, Andrew~L, Daly, Raymond~E, Pham, Peter~T, Huang, Dan, Ng, Andrew~Y, and
  Potts, Christopher.
\newblock {Learning Word Vectors for Sentiment Analysis}.
\newblock In \emph{Proceedings of ACL}, 2011.

\bibitem[Mikolov et~al.(2013)Mikolov, Chen, Corrado, and Dean]{Mikolov2013a}
Mikolov, Tomas, Chen, Kai, Corrado, Greg, and Dean, Jeffrey.
\newblock {Distributed Representations of Words and Phrases and their
  Compositionality}.
\newblock In \emph{NIPS}, 2013.

\bibitem[Schwenk(2012)]{Schwenk2012}
Schwenk, Holger.
\newblock {Continuous Space Translation Models for Phrase-Based Statistical
  Machine Translation}.
\newblock In \emph{International Conference on Computational Linguistics}, pp.\
   1071--1080, 2012.

\bibitem[Simonyan et~al.(2013)Simonyan, Vedaldi, and Zisserman]{Simonyan2013}
Simonyan, Karen, Vedaldi, Andrea, and Zisserman, Andrew.
\newblock {Deep Inside Convolutional Networks: Visualising Image Classification
  Models and Saliency Maps}.
\newblock Technical report, 2013.

\bibitem[Socher et~al.(2011)Socher, Pennington, Huang, Ng, and
  Manning]{Socher2011}
Socher, Richard, Pennington, Jeffrey, Huang, Eric~H, Ng, Andrew~Y, and Manning,
  Christopher~D.
\newblock {Semi-Supervised Recursive Autoencoders for Predicting Sentiment
  Distributions}.
\newblock In \emph{Conference on Empirical Methods in Natural Language
  Processing}, number~i, 2011.

\bibitem[Socher et~al.(2012)Socher, Huval, Manning, and Ng]{Socher2012}
Socher, Richard, Huval, Brody, Manning, Christopher~D, and Ng, Andrew~Y.
\newblock {Semantic Compositionality through Recursive Matrix-Vector Spaces}.
\newblock In \emph{Conference on Empirical Methods in Natural Language
  Processing}, 2012.

\bibitem[Sutskever et~al.(2014)Sutskever, Vinyals, and Le]{Sutskever2014}
Sutskever, Ilya, Vinyals, Oriol, and Le, Quoc V.~V.
\newblock Sequence to sequence learning with neural networks.
\newblock In \emph{Advances in Neural Information Processing Systems 27}, pp.\
  3104--3112, 2014.

\bibitem[Weston et~al.(2014)Weston, Chopra, and Bordes]{WestonCB14}
Weston, Jason, Chopra, Sumit, and Bordes, Antoine.
\newblock Memory networks.
\newblock \emph{CoRR}, abs/1410.3916, 2014.

\bibitem[Yu et~al.(2014)Yu, Hermann, Blunsom, and Pulman]{Yu2014}
Yu, Lei, Hermann, Karl~Moritz, Blunsom, Phil, and Pulman, Stephen.
\newblock Deep learning for answer sentence selection.
\newblock In \emph{NIPS Deep Learning Workshop}, 2014.

\bibitem[Zeiler \& Fergus(2012)Zeiler and Fergus]{Zeiler2012}
Zeiler, Matthew~D and Fergus, Rob.
\newblock {Visualizing and Understanding Convolutional Networks}.
\newblock Technical report, 2012.

\bibitem[Zeiler et~al.(2010)Zeiler, Krishnan, Taylor, and Fergus]{Zeiler2010}
Zeiler, Matthew~D, Krishnan, Dilip, Taylor, Graham~W, and Fergus, Rob.
\newblock {Deconvolutional Networks}.
\newblock In \emph{Computer Vision and Pattern Recognition}, 2010.

\bibitem[Zeiler et~al.(2011)Zeiler, Taylor, and Fergus]{Zeiler2011}
Zeiler, Matthew~D, Taylor, Graham~W, and Fergus, Rob.
\newblock {Adaptive Deconvolutional Networks for Mid and High Level Feature
  Learning}.
\newblock In \emph{International Conference on Computer Vision}, 2011.

\end{thebibliography}
\bibliographystyle{iclr2015}
}

\end{document}